\documentclass[runningheads]{llncs}
\usepackage[T1]{fontenc}
\usepackage{graphicx,verbatim}
\usepackage{booktabs}
\usepackage{multirow}
\usepackage[table]{xcolor}
\usepackage{amsmath}
\usepackage{caption}

\begin{document}

\title{NeRD: Neuro-Symbolic Rule Distillation for Efficient Ontology-Grounded Chain-of-Thought in Medical Image Diagnosis}
\titlerunning{NeRD}

\author{Hongxi Yang\inst{1,2} \and
Yiwen Jiang\inst{2,3}$^{\star}$ \and
Siyuan Yan\inst{1,2} \and
Jamie Chow\inst{4} \and
Eunis Li\inst{4} \and
Charlotte Poon\inst{4} \and
Stephanie Fong\inst{1,2} \and
Xiangyu Zhao\inst{1,2} \and
Deval Mehta\inst{1,5} \and
Yasmeen George\inst{1,2} \and
Zongyuan Ge\inst{1,2}\thanks{Corresponding author.}}

\authorrunning{Hongxi Yang et al.}
%
\institute{Department of Data Science \& AI, Faculty of Information Technology, Monash University, Melbourne, Australia 
\and
AIM for Health Lab, Faculty of Information Technology, Monash University, Melbourne, Australia\\
\and
Faculty of Engineering, Monash University, Melbourne, Australia\\
\and
Faculty of Medicine, The Chinese University of Hong Kong, Hong Kong, China\\
\and
School of Computing Technologies, RMIT University, Melbourne, Australia
}


\maketitle              
\begin{abstract}

Interpretability is essential for trustworthy medical image diagnosis. However, existing concept-driven interpretable methods have key limitations: Concept Bottleneck Models (CBMs) require scoring all predefined concepts at inference time and for manual intervention, imposing a substantial burden on clinicians, while rationale-based generative approaches often select concepts by class discriminability, which can drift from diagnostic ontologies. To address these issues, we propose \textbf{Neuro-Symbolic Rule Distillation (NeRD)}, a framework that produces efficient, ontology-grounded reasoning chains that are sufficient yet non-redundant, without manually crafting diagnostic rules. Experiments on two skin datasets demonstrate strong diagnostic performance and interpretability, and blinded expert evaluation confirms the clinical plausibility of NeRD rationales. Our method further enables a first expert-in-the-loop study for Multimodal Chain-of-Thought-based diagnosis, achieving efficient and effective concept-level intervention.

\keywords{Explainable AI \and Chain of Thought \and Concept\and Skin disease diagnosis.}

\end{abstract}
\section{Introduction}

\begin{figure}[t]
\centering
\includegraphics[width=\textwidth]{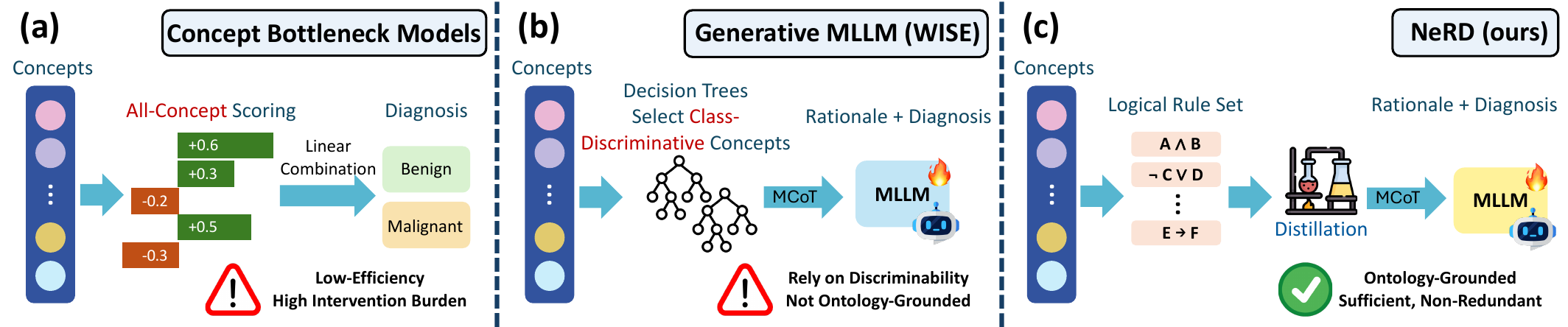}
\caption{Comparison of concept-driven interpretable methods. (a) CBM (discriminative) predicts the diagnosis via a linear combination of all concept scores. (b) WISE (generative) uses decision-tree guidance to select class-discriminative concepts and reformulate them into MCoTs. (c) Ours distills ontology-grounded diagnostic MCoTs from induced logical rule set.} \label{fig1}
\end{figure}

Real-world clinical decision-making relies on inherently interpretable and structured reasoning, where visual evidence is evaluated and explicitly mapped to established diagnostic criteria~\cite{croskerry2009clinical,yang2024deep,10.1007/978-3-031-96625-5_18}. Recently, deep learning (DL) models have shown strong potential in medical image diagnosis \cite{yan2026visionlanguagefoundationmodelzeroshot,yang2026multimodal,yang2026semantic}, yet their opaque nature limits clinical trust and safe adoption in practice~\cite{papernot2017practical}.
Existing interpretability methods for DL largely fall into two paradigms~\cite{linardatos2020explainable}: discriminative concept based explanations such as Concept Bottleneck Models (CBMs)~\cite{jiang-etal-2025-enhancing,koh2020concept}, and generative rationale based stepwise reasoning such as Multimodal Chain of Thought (MCoT)~\cite{zhang2024visually,zhang2023multimodal} for Multimodal Large Language Models (MLLMs)~\cite{yin2024survey}. Despite their potential, CBMs require evaluating all predefined concepts at inference time, whereas MCoT depends on substantial expert authored rationales, both introducing computational and annotation burden. Bridging these paradigms while addressing limitations, WISE~\cite{jiang2025wise} is the first to automate MCoT generation from the bottleneck layer of CBMs. It leverages an ensemble of decision trees to selectively activate class-discriminative concepts, yielding concept-grounded rationales while avoiding the exhaustive, all-concept scoring pipeline of CBMs (Fig.~\ref{fig1}(a,b)).

However, concept selection guided purely by class discriminability can diverge from established diagnostic ontologies, which leads to several limitations of WISE for medical image diagnosis. First, it may favor statistically predictive but clinically irrelevant shortcuts, selecting concepts that fall outside the ontology's criterion-based decision pathway and thereby bypassing the guideline-encoded diagnostic logic. Second, the generated MCoTs lack systematic validation by human experts, raising uncertainty about their alignment with clinical reasoning and limiting confidence in deployment. Third, CBMs natively support concept-level intervention, allowing clinicians to correct intermediate concept predictions and propagate these corrections to the final diagnosis~\cite{yan2023towards}. However,  comparable intervention in generative MLLM reasoning remains largely unexplored, leaving a critical gap for human-in-the-loop clinical workflows.

To address these limitations, we propose \textbf{Neuro-symbolic Rule Distillation (NeRD)}, a framework that produces efficient, ontology-grounded MCoT rationales for interpretable medical image diagnosis (Fig.~\ref{fig1}(c)). Rather than prioritizing class discriminability, NeRD first induces explicit logical diagnostic rules over clinical concepts in a data-driven manner, without manual rule engineering. It then distills the activated rules into compact, case-specific reasoning chains through rule selection, logic simplification, and concept grounding, thereby removing redundant concepts and yielding concise, ontology-aligned evidence for each prediction. \textbf{Our contributions are threefold}: (1) We propose NeRD, a novel framework that bridges rule-based concept interpretability with MCoT reasoning for medical image diagnosis. (2) We design a blinded human expert study to assess the clinical plausibility, showing that NeRD rationales best support expert diagnostic reasoning. (3) We present the first study of concept intervention in MCoT-based diagnosis, demonstrating that NeRD enables both efficient and effective concept-level intervention.

\section{Method}

\begin{figure}[t]
\centering
\includegraphics[width=\textwidth]{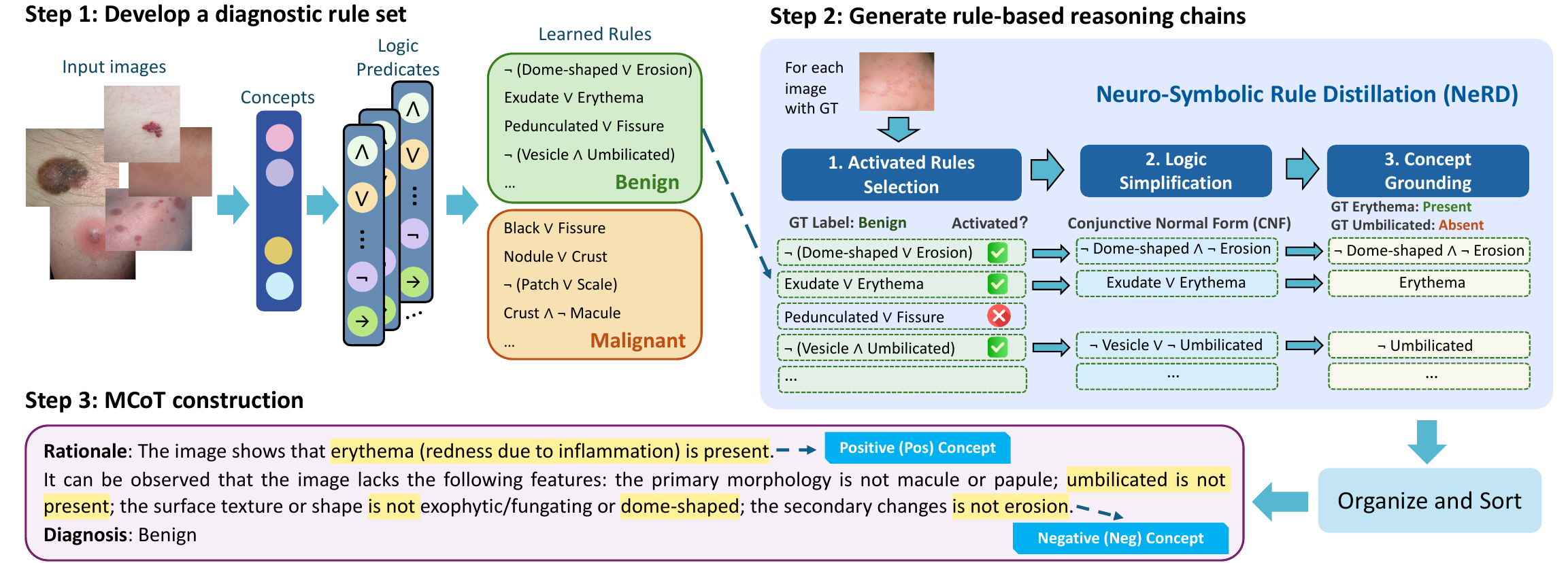}
\caption{Framework of our method. Step 1: Induce a diagnostic rule set from concepts. Step 2: Execute Neuro-Symbolic Rule Distillation (NeRD) to select, simplify, and ground rules. Step 3: Organize the reasoning chains into MCoT.} \label{fig2}
\end{figure}

\subsubsection{Task Formulation.}
Let $\mathcal{S}=\{(x_i,\mathbf{c}_i,y_i)\}_{i=1}^{N}$ denote a dataset, where $x_i$ is an input image, $y_i\in\mathcal{Y}$ is the diagnostic label, and $\mathbf{c}_i=[c_i^{1},c_i^{2},\dots,c_i^{K}]$ represents annotations for $K$ concepts. Our goal is to derive, for each $x_i$, a compact textual MCoT that selects a small subset of concepts and verbalizes them into a reasoning chain to fine-tune MLLMs. As illustrated in Fig.~\ref{fig2}, our framework operates in three steps. First, we begin by inducing a diagnostic rule set to capture the underlying logic between concept pairs. Next, NeRD distills these rules into compact, case-specific reasoning chains through rule activation, logic simplification, and concept grounding. Finally, the refined chains are structured into MCoTs.

\subsection{Diagnostic Rule Set Construction}

We induce a diagnostic rule set $\mathcal{R}$ using LogicCBM~\cite{vemuri2025logiccbms} in Step 1 of Fig.~\ref{fig2}. LogicCBM learns differentiable logical gates (e.g., $\wedge$, $\vee$, $\neg$) to compose pairs of concept predicates into human-interpretable rules such as $r_j = c_i^a \wedge \neg c_i^b$, defined over $\mathbf{c}_i$. Each derived rule $r_j \in \mathcal{R}$ is associated with a class-wise weight vector. We assign $r_j$ to a diagnosis $y\in\mathcal{Y}$ based on these weights: if signs differ across classes, $r_j$ is assigned to the class with a positive weight; otherwise, it is assigned to the class with the largest weight. This yields a class-conditioned partition $\{\mathcal{R}^{y}\}_{y \in \mathcal{Y}}$, where each subset $\mathcal{R}^{y}$ collects rules attributed to diagnosis $y$.

\subsection{Neuro-Symbolic Rule Distillation}

\subsubsection{Activated Rule Selection.}

In Step 2 of Fig.~\ref{fig2}, for each training sample $(x_i,\mathbf{c}_i,y_i)$, we restrict candidate rules to the class-specific subset $\mathcal{R}^{y_i}$. We then perform rule-level selection by applying each $r_j \in \mathcal{R}^{y_i}$ to the concept annotation $\mathbf{c}_i$, and retain only the activated rules (i.e., those whose logical predicates are satisfied). This activation-based screening prunes rules that do not explain the current case, yielding a diagnostically aligned rule set for subsequent refinement.

\subsubsection{Logic Simplification.}

Since activated rules in $\mathcal{R}^{y_i}$ are selected independently without cross-rule interaction, they may contain overlapping or redundant conditions. To obtain a compact representation, we combine all selected rules via conjunction and convert the resulting expression into Conjunctive Normal Form (CNF). Specifically, an selected rule $r_j$ is rewritten as a conjunction of disjunctive clauses, formulated as $r_j^{\text{CNF}} = \bigwedge_{l=1}^{L_j} D_{j,l}$, where $D_{j,l} = \bigvee_{k} \ell_k$. Here, each literal $\ell_k$ corresponds to either a positive concept assertion ($c^k$) or its negation ($\neg c^k$). This conversion utilizes standard logical equivalences, including De Morgan's laws~\cite{de1847formal}. After applying this transformation across all activated rules, we pool the resulting disjunctive clauses and eliminate duplicates, yielding a simplified, unique clause set $\mathcal{C}_i = \{D_1, D_2, \dots, D_P\}$.

\subsubsection{Concept Grounding.}
$\mathcal{C}_i$ may still contain residual disjunctions (uncertainties) that need to be resolved for the specific sample. To ground this logic, we evaluate every literal $\ell_k$ against the concept annotations $\mathbf{c}_i$. For each clause $D_l$, we extract only the literals that evaluate to True, forming a set of verified facts: $D_l' = \{\ell \in D_l \mid \ell(\mathbf{c}_i) = \text{True}\}$. For instance, given $D_l = \text{Exudate} \vee \text{Erythema}$, if $\text{Erythema}$ is present and $\text{Exudate}$ is absent in $\mathbf{c}_i$, the clause reduces to $\{\text{Erythema}\}$. We then aggregate these verified literals into a unified grounded set $\mathcal{G}_i = \bigcup_{l=1}^{P} D_l'$. At this stage, $\mathcal{G}_i$ consists purely of conjunctive assertions.

Additionally, some clinical concepts form mutually exclusive groups (e.g., $\text{pigment network} \in \{\text{absent}, \text{atypical}, \allowbreak \text{typical}\}$). We leverage this structural prior to prune redundant negations in $\mathcal{G}_i$. Specifically, if $\mathcal{G}_i$ contains a positive assertion $c^v$ (e.g., typical) and a negation $\neg c^t$ (e.g., $\neg$atypical) belonging to the same group, $\neg c^t$ is safely discarded; the presence of $c^v$ implicitly guarantees the absence of all other states in that concept group. Conversely, if all concepts in a group except one are negated in $\mathcal{G}_i$ (e.g., both $\neg$typical and $\neg$atypical are present), we logically infer the remaining concept (i.e., absent) as a positive assertion and update the set accordingly.

\subsection{MCoT Construction.}
In Step 3 of Fig.~\ref{fig2}, we construct MCoT rationale from the grounded set $\mathcal{G}_i$. We first partition $\mathcal{G}_i$ into supportive (positive) concepts $\mathcal{G}_i^{+}$ and refutational (negative) concepts $\mathcal{G}_i^{-}$. To produce a coherent and clinically intuitive narrative, concepts within each partition are prioritized based on their class-conditional statistics estimated on the training set. Concretely, concepts in $\mathcal{G}_i^{+}$ are ranked by decreasing prevalence under the predicted label, whereas concepts in $\mathcal{G}_i^{-}$ are ranked by the label-conditional probability of being absent. To preserve structural cohesion, we further group concepts by clinical attribute category (e.g., primary morphology, texture) and order concepts within each group accordingly. Finally, we render the rationale with a fixed template: (i) report the salient positive findings, (ii) enumerate clinically relevant absent features, and (iii) conclude with the diagnostic prediction.

\section{Experiments}
\subsection{Experimental Datasets}

We evaluate NeRD on two publicly available skin image datasets sourced from MAKE~\cite{11446061,yan2025make}: Derm7pt~\cite{kawahara2018seven} and Fitzpatrick17k (F17k)~\cite{groh2021evaluating}. Derm7pt provides annotations for seven categorical clinical concept groups, with mutually exclusive values (e.g., $\text{pigment network} \in \{\text{absent}, \text{atypical}, \allowbreak \text{typical}\}$), yielding a 28-dimensional one-hot concept vector. We follow the official split (413/203/395 for train/val/test) and perform melanoma vs.\ non-melanoma classification. F17k includes 32 independently annotated binary concepts. Following~\cite{pang2024integrating}, we formulate the diagnostic task as benign vs.\ malignant and perform a stratified split into 2253/322/644 cases for train/val/test.

\subsection{Experiment 1: Human Expert Evaluation of Clinical Plausibility}

\begin{figure}[t]
\centering
\includegraphics[width=\textwidth]{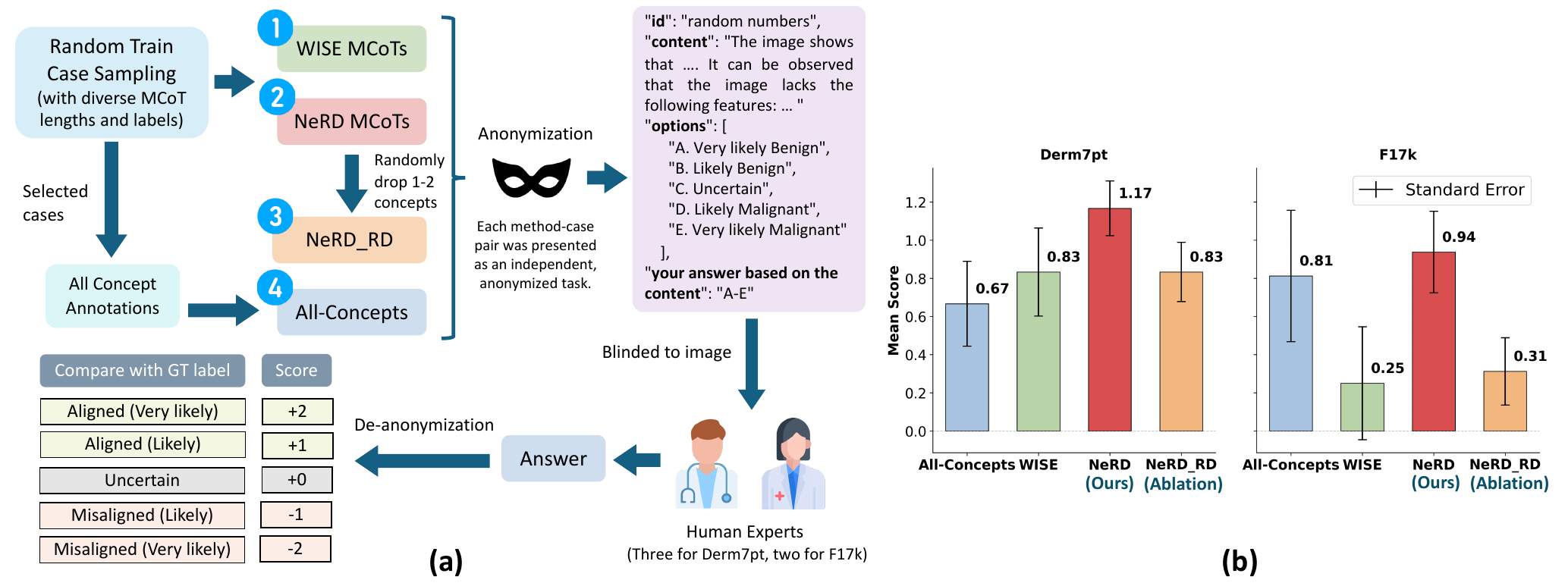}
\caption{Human expert evaluation overview. (a) Workflow of the human expert evaluation. We sample training cases and rationales from four methods. The rationales are anonymized and presented to experts, who are asked to select a diagnosis based solely on the rationale content. (b) Evaluation results. Mean scores ($\pm$ standard error) across evaluators on Derm7pt and F17k. Higher scores indicate stronger alignment between expert judgments and diagnostic labels.} \label{fig3}
\end{figure}

\subsubsection{Evaluation Method.}
To assess the clinical plausibility of the reasoning chains, we conduct a blinded evaluation with dermatology-trained experts (three for Derm7pt, two for F17k) experienced in concept-based skin lesion diagnosis. As illustrated in Fig.~\ref{fig3} (a), for each evaluator we randomly sample 8 training cases, spanning diverse diagnostic labels and chain lengths; each case yields 4 distinct text-only rationales from All-Concepts (verbalizing all annotated concepts), WISE~\cite{jiang2025wise}, NeRD (ours), and NeRD-RD (an ablation that randomly removes 1-2 concepts). To prevent cueing, each rationale is detached from its originating case and method and presented as an independent item, all items are globally shuffled across cases and methods. Evaluators assign a 5-point ordinal likelihood from A (very likely benign/non-melanoma) to E (very likely malignant/melanoma). We score responses by agreement with the ground-truth label: correct predictions receive $+2$ (A/E) or $+1$ (B/D), incorrect predictions receive $-2$ or $-1$ with the same confidence mapping, and uncertain responses (C) receive $0$. Scores are averaged over all items to obtain a mean score per method.

\subsubsection{Main Results.}
Fig.~\ref{fig3}(b) summarizes the results of the human expert study. NeRD attains the highest mean score on both datasets (Derm7pt: 1.17; F17k: 0.94), surpassing WISE (0.83; 0.25) and All-Concepts (0.67; 0.81). 
These results suggest that NeRD mitigates the redundancy in All-Concepts, where redundant or irrelevant details can dilute salient evidence, potentially increasing cognitive load and reducing diagnostic confidence. NeRD also selects concepts that are more clinically aligned than WISE. To further assess the contribution of NeRD’s selection, we perform an ablation by randomly removing one to two concepts (NeRD-RD). This drives a marked performance drop (Derm7pt: 1.17 $\rightarrow$ 0.83; F17k: 0.94 $\rightarrow$ 0.31), confirming the importance and non-redundancy of NeRD's selected concepts for reaching correct conclusions without image access.

\subsection{Experiment 2: Fine-Tuned MLLMs for Interpretable Diagnosis}

\begin{table}[t]
\centering
\caption{Comparison of diagnostic and concept prediction performance. ACC (\%): diagnostic accuracy; F1 (\%): diagnostic F1 score; Intp. (\%): interpretability, quantified by concept accuracy (at the bottleneck layer or along the MCoT).}
\label{tab:results}
\setlength{\tabcolsep}{1.5pt}
\begin{tabular}{l|l|ccc|ccc}
\toprule
\multicolumn{2}{c|}{\multirow{2}{*}{Method}} & \multicolumn{3}{c|}{Derm7pt} & \multicolumn{3}{c}{F17k} \\
\cmidrule(lr){3-5} \cmidrule(lr){6-8}
\multicolumn{2}{c|}{} & ACC & F1 & Intp. & ACC & F1 & Intp. \\
\midrule
Black-box & InceptionV3 & \textbf{81.77} & 55.00 & - & 88.04 & 54.97 & - \\
\cline{1-8}
\multirow{4}{*}{\shortstack[l]{Discriminative\\concept-based}} & Joint CBM & 79.49 & 50.91 & 75.01 & 88.35 & 52.83 & 79.71 \\
 & Sequential CBM & 78.99 & 34.65 & \textbf{78.80} & 76.86 & 13.87 & 91.59 \\
 & Joint LogicCBM & 79.75 & \underline{58.33} & 62.75 & 87.73 & 55.87 & 69.01 \\
 & Sequential LogicCBM & 77.97 & 54.45 & 64.76 & \underline{88.51} & \underline{57.95} & 71.77 \\
\cline{1-8}
\multirow{3}{*}{\shortstack[l]{Generative\\reasoning}} & Zero-shot & 77.22 & 32.84 & 56.78 & 84.63 & 34.44 & 88.62 \\
 & WISE & 77.72 & 43.59 & 57.06 & 86.65 & 56.12 & 85.90 \\
 & \cellcolor{gray!25}NeRD (Ours) & \cellcolor{gray!25} \underline{80.76} & \cellcolor{gray!25} \textbf{60.00} & \cellcolor{gray!25} \underline{76.76} & \cellcolor{gray!25} \textbf{88.66} & \cellcolor{gray!25} \textbf{61.70} & \cellcolor{gray!25} \textbf{94.10} \\
\bottomrule
\end{tabular}
\end{table}

\subsubsection{Baselines and Implementation Details.}

We compare NeRD against three categories of baselines: (1) Black-box models. An InceptionV3~\cite{szegedy2016rethinking} model trained end-to-end from images to diagnostic labels, without concept supervision. (2) Discriminative concept-based models. Sequential CBM~\cite{koh2020concept}, Joint CBM~\cite{koh2020concept}, as well as LogicCBM~\cite{vemuri2025logiccbms} in both joint and sequential training variants, all using InceptionV3 as the backbone. (3) Generative reasoning models. Zero-shot prompting with Qwen3-VL-8B-Instruct~\cite{bai2025qwen3} using the predefined concept set and task description to elicit concept-based rationales without any fine-tuning; and WISE~\cite{jiang2025wise}, which generates MCoTs via decision-tree–guided reformulation. For MLLM fine-tuning, we adopt Qwen3-VL-8B-Instruct as the backbone and perform parameter-efficient adaptation with LoRA~\cite{hu2022lora} via LLaMA-Factory~\cite{zheng2024llamafactory}. Fine-tuning is conducted for 8 epochs on a single NVIDIA RTX 4090. All models are trained on the training split, selected by validation diagnostic accuracy, and evaluated on the test split.

\subsubsection{Evaluation Method.}

We evaluate diagnostic performance using accuracy and F1 score. To measure interpretability, we compute concept accuracy over the concepts explicitly mentioned in the generated MCoT. Since Zero-shot MCoT outputs are not emitted in a fixed schema, we use GPT-5 Mini~\cite{singh2025openai} to parse the predicted concepts and diagnosis from free-form text for evaluation.

\begin{figure}[t]
\centering
\includegraphics[width=\textwidth]{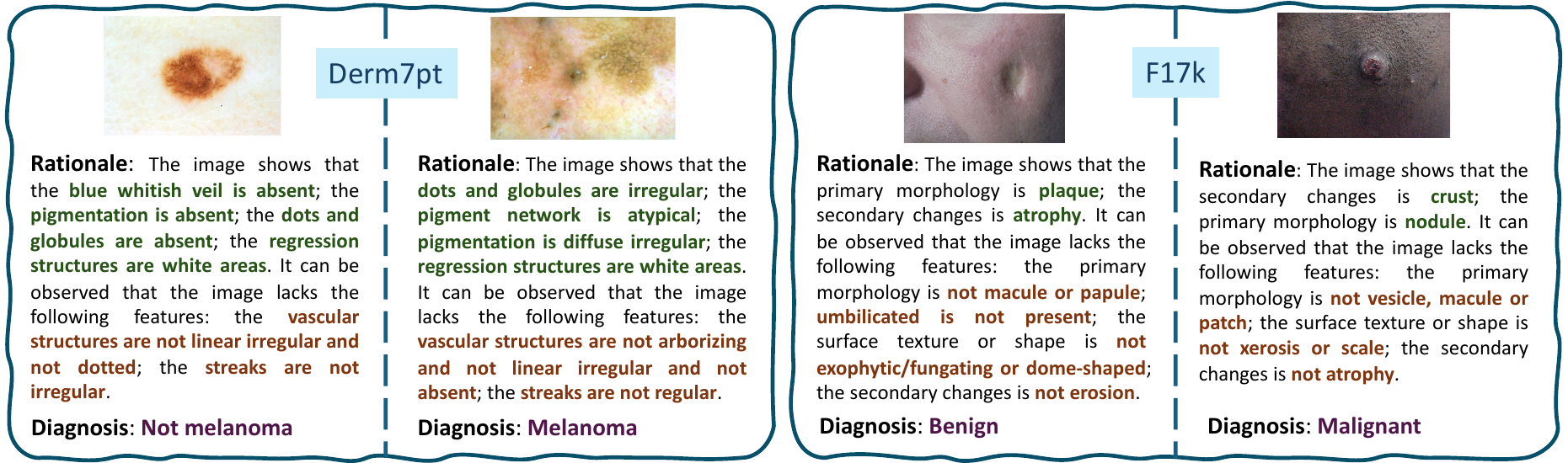}
\caption{NeRD generated MCoTs.} \label{fig4}
\end{figure}

\begin{table}[t]
\centering
\begin{minipage}[t]{0.45\textwidth}
\centering

\caption{Average concept count. Pos / Neg: present / absent per case (CBM); supportive / refutational per MCoT (WISE and NeRD).}
\label{tab:concept count}
\setlength{\tabcolsep}{0.7pt}
\begin{tabular}{l|ccc|ccc}
\toprule
\multirow{2}{*}{Method} & \multicolumn{3}{c|}{Derm7pt} & \multicolumn{3}{c}{F17k} \\
\cmidrule(lr){2-4} \cmidrule(lr){5-7}
 & Pos & Neg & Total & Pos & Neg & Total \\
\midrule
CBM & 7 & 21 & 28 & 3 & 29 & 32 \\
WISE & 3 & 1 & 4 & 2 & 11 & 13 \\
\rowcolor{gray!25}
NeRD & 3 & 3 & 6 & 1 & 5 & 6 \\
\bottomrule
\end{tabular}
\end{minipage}
\hfill
\begin{minipage}[t]{0.53\textwidth}
\centering
\caption{Intervention results on Derm7pt. N: intervened samples; Concepts: average corrected concepts per sample; Before / After: diagnostic accuracy (\%) pre / post intervention; Change: accuracy gain.}
\label{tab:intervention}
\setlength{\tabcolsep}{0.7pt}
\begin{tabular}{l|c|c|c|c|c}
\toprule
 \multirow{1}{*}{Method} & N & Concept & Before & After & Change \\
\midrule
CBM & 82 & 6.56 & 78.99 & 83.04 & +4.05 \\
\rowcolor{gray!25}
NeRD & 73 & 2.26 & 80.76 & 85.82 & +5.06 \\
\bottomrule
\end{tabular}
\end{minipage}
\end{table}

\subsubsection{Main Results.}
Table~\ref{tab:results} reports diagnostic performance and concept prediction accuracy on Derm7pt and F17k datasets. Among generative methods, NeRD achieves the strongest overall results: 80.76\% ACC / 60.00\% F1 on Derm7pt and 88.66\% ACC / 61.70\% F1 on F17k, outperforming WISE and the zero-shot baseline. On Derm7pt, NeRD attains the best ACC among interpretable methods and ranks second overall, trailing only the black-box InceptionV3; on F17k, it yields the highest ACC across all compared methods. For concept prediction, NeRD substantially improves concept accuracy over WISE on both datasets (76.76\% vs.\ 57.06\% on Derm7pt; 94.10\% vs.\ 85.90\% on F17k). Fig.~\ref{fig4} provides qualitative examples of the generated MCoTs. Table~\ref{tab:concept count} further shows that NeRD uses 6 concepts per MCoT for both datasets, whereas CBM-style baselines evaluate all predefined concepts (28 for Derm7pt; 32 for F17k). This indicates that NeRD can preserve competitive interpretability while producing compact, non-exhaustive reasoning chains. Notably, on F17k, WISE yields more negative concepts per chain (11 vs.\ 5), consistent with its tendency to continue exploring additional discriminative cues when the current concept set fails to separate classes.

\subsection{Experiment 3: Clinician Intervention on Reasoning Concepts}

\subsubsection{Intervention Setup and Evaluation.}
We evaluate the efficiency and efficacy of concept intervention on Derm7pt by comparing NeRD with a Sequential CBM~\cite{koh2020concept}. For both methods, we intervene only on test cases where the model (i) predicts an incorrect diagnosis and (ii) makes at least one concept prediction error. This protocol mirrors a realistic workflow in which clinicians review and correct concept-level mistakes only for flagged, high-risk cases. For NeRD, we locate the incorrect concept statements in the generated MCoTs, replace them with the ground-truth concept annotations, and prepend the corrected rationale to prompt the model to regenerate the diagnosis. For the Sequential CBM, we overwrite the incorrectly predicted concept variables with their annotations and forward-propagate the corrected concept vector through the concept-to-label predictor. No parameters are updated during intervention. We report post-intervention diagnostic performance, along with the intervention cost measured by the number of corrected concepts.

\subsubsection{Main Results.}
Table~\ref{tab:intervention} summarizes intervention results on Derm7pt. Across 73 intervened cases, NeRD requires only 2.26 concept corrections  per sample on average while improving accuracy by 5.06\% (80.76\%$\rightarrow$85.82\%). By comparison, Sequential CBM requires 6.56 corrections per sample, yet yielding a smaller gain of 4.05\% (78.99\%$\rightarrow$83.04\%) across 82 intervened cases. Overall, NeRD demonstrates higher intervention-efficient refinement by enabling larger diagnostic improvements with substantially fewer concept edits.

\section{Conclusion}

We propose NeRD, a neuro-symbolic framework that bridges rule-based concept interpretability with MCoT reasoning for medical image diagnosis. By selecting concepts based on diagnostic necessity, NeRD distills induced logical rules into compact, criterion-aligned reasoning chains. Experiments on two skin image datasets demonstrate strong diagnostic performance and concept fidelity.  Blinded expert evaluation further confirms the clinical plausibility of NeRD rationales, while intervention experiments demonstrate efficient concept-level correction with fewer edits and larger diagnostic gains.

\subsubsection*{Acknowledgments.} This work was supported by Australian Government under Grant CRCPXIII000022.

\bibliographystyle{splncs04}
\bibliography{ref}

\end{document}